\pdfoutput=1

\documentclass[11pt]{article}

\usepackage[]{EMNLP2022}

\usepackage{times}
\usepackage{latexsym}

\usepackage[T1]{fontenc}

\usepackage[utf8]{inputenc}

\usepackage{microtype}
\usepackage{graphicx}
\usepackage{caption}
\usepackage{subcaption}
\usepackage{booktabs}
\usepackage{multirow}
\usepackage[linesnumbered,ruled]{algorithm2e}
\usepackage{amsmath,amssymb}

\newcommand{\nop}[1]{}
\usepackage{soul}
\usepackage[normalem]{ulem}

\newif\ifdebugdoc\debugdocfalse

\ifdebugdoc
\newcommand{\fromJ}[1]{\textcolor{blue}{J: #1}}
\newcommand{\anjie}[1]{\textcolor{purple}{Anjie: #1}}
\newcommand{\todo}[1]{\textcolor{red}{#1}}
\else
\newcommand{\fromJ}[1]{}
\newcommand{\anjie}[1]{}
\newcommand{\todo}[1]{}
\fi

\usepackage{graphicx}

\title{Reinforced Question Rewriting for Conversational Question Answering}

 \author{Zhiyu Chen, Jie Zhao, Anjie Fang, Besnik Fetahu, Oleg Rokhlenko, Shervin Malmasi  \\ 
            Amazon.com, Inc., Seattle, WA, USA \\
            \texttt{\{zhiyuche,zhaozjie,njfn,besnikf,olegro,malmasi\}@amazon.com}}

\begin{document}
\maketitle
\begin{abstract}

Conversational Question Answering (CQA) aims to answer questions contained within dialogues, which are not easily interpretable without context.
Developing a model to rewrite conversational questions into self-contained ones is an emerging solution in industry settings as it allows using existing single-turn QA systems to avoid training a CQA model from scratch. Previous work trains rewriting models using human rewrites as supervision. However, such objectives are disconnected with QA models and therefore more human-like rewrites do not guarantee better QA performance.
.

In this paper we propose using QA feedback to supervise the rewriting model with reinforcement learning. %
Experiments show that our approach can effectively improve QA performance over baselines for both extractive and retrieval QA. Furthermore, human evaluation shows that our method can generate more accurate and detailed rewrites when compared to human annotations.

\end{abstract} 
\section{Introduction}\label{sec:intro}

Interacting through conversations is a natural information-seeking procedure for humans, therefore it is important for AI assistants like Apple Siri and Amazon Alexa to enable and improve such experiences. In recent years Conversational Question Answering (CQA) has gained more attention, where a user can ask a series of related questions and ideally obtain answers that leverage the conversational context. 

Different from widely-studied question answering (QA) tasks that happen in single-turn \cite{rajpurkar-etal-2016-squad,rajpurkar-etal-2018-know,tay2018hyperbolic,tang2019multi}, the interpretation of conversational questions in CQA depends on questions and answers from previous turns. 

\begin{figure}[t]
\centering
\includegraphics[width=0.4\textwidth]{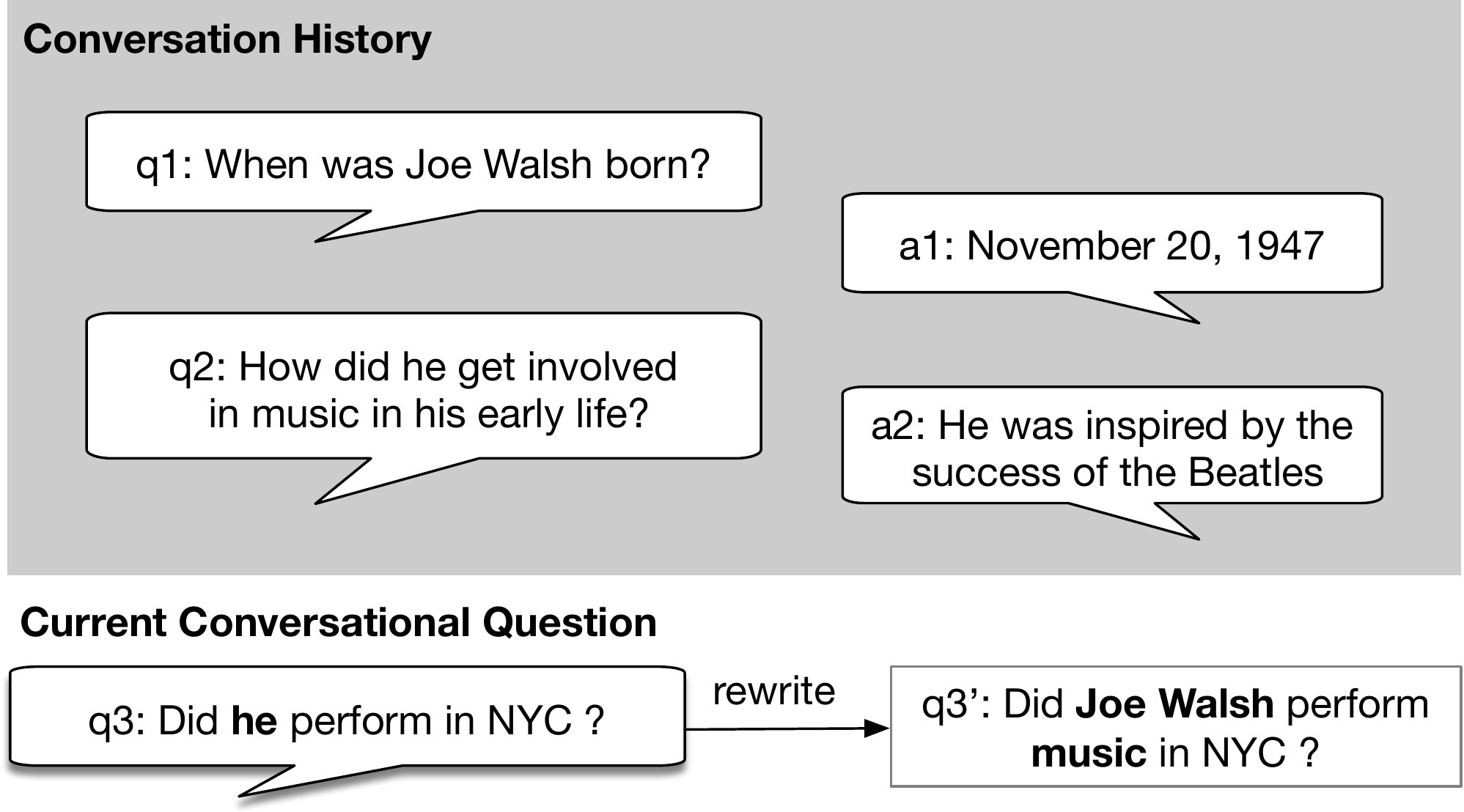}
\caption[]{A conversational question rewriting example. }
\label{fig:conv_exp}
\end{figure}

Previous approaches to CQA usually train new models from scratch, which can achieve promising results but also are expensive in terms of obtaining domain-specific training data. 
In industry settings, there are many single-turn QA models deployed. Training new CQA models with additional annotations to replace each existing single-turn QA model is expensive, and generally not feasible.
Moreover, discarding existing single-turn models and datasets is impractical, and studying how to reuse these existing resources to tackle CQA  merits attention.

Existing approaches to this task, called Conversational Question Rewriting (CQR),  often train sequence-to-sequence models supervised by human rewrites to generate self-contained questions~\citep{ren2018conversational,vakulenko2021question}. Such methods have several limitations. First, the CQR training objective is disconnected from CQA performance. The annotation process of existing rewriting datasets has no knowledge of the QA systems, and more human-like rewrites do not guarantee better CQA performance. Second, the rewriting model does not take into account the feedback from downstream QA systems. In industry settings, multiple single-turn QA systems trained with different datasets serve in the backend. It is impractical to replace them with new CQA models, and we argue that their output can still be used as signals to help train rewriting models.

To overcome these limitations, we propose an effective CQR approach upon the recent success of Reinforcement Learning (RL) techniques for text generation \cite{rennie2017self}. RL enables flexible ways to incorporate training objectives in the form of reward functions. We systematically analyze different rewards and their effectiveness in terms of final QA performance, as well as the quality of the question rewrites (i.e. the question still has to be understandable and interpretable by humans). 
To optimize QA performance, we propose various QA rewards to measure the likelihood of a question yielding a better answer.
In comparison with the QA rewards, we also propose to use the same RL approach with question rewriting (QR) rewards reflecting the similarity between a model-generated question and the human's ground-truth.

We summarize our contributions as follows:
\begin{itemize}
\setlength\itemsep{-0.3em}
    \item To the best of our knowledge, we are the first to study how to incorporate QA signals to improve CQR using RL.
    \item We systematically propose and compare using different training signals as rewards. 
    \item We conduct experiments on two CQA tasks to show our approach is effective.
    \item A user study shows that our method can generate more accurate and detailed rewrites when compared to human annotations.
\end{itemize}

\section{Related Work}
\noindent \textbf{Conversational Question Answering.}
Recently, conversational QA has been studied which presents new challenges for QA models such as being able to resolve conversational dependencies so that a conversational question can be interpreted by QA models with conversational context. 
QuAC~\cite{elgohary-etal-2019-unpack} and CoQA~\cite{reddy2019coqa} are two datasets for extractive CQA where answers are extracted from passages.  CAsT-19~\cite{Dalton2020TRECC2} is a benchmark for retrieval CQA and the target is to return relevant passages given a question. QReCC~\cite{anantha-etal-2021-open} combines retrieval and extractive CQA where the answers are extracted from passages returned by a retrieval component. \citet{kim-etal-2021-learn} propose to train the CQA model and rewriter simultaneously, which is impractical for industry setting. A directly related work to ours is \citet{vakulenko2021question} which proposes to rewrite questions for CQA. However, they do not consider taking the QA feedback into the CQR training which is studied in our work.

\noindent \textbf{RL for Nature Language Generation.}
Reinforcement learning methods have been widely applied for various language generation tasks. \citet{li-etal-2016-deep} propose to apply deep reinforcement learning in dialogue generation to model future rewards related to conversational properties, such as informativeness, coherence and ease of answering. \citet{DBLP:journals/corr/RanzatoCAZ15} propose Mixed Incremental Cross-Entropy Reinforce~(MIXER) for sequence prediction to directly optimize the metrics used at test time, such as BLEU or ROUGE. They show MIXER outperforms several strong baselines for greedy generation on text summarization, image caption and machine translation. 
\citet{nogueira2017task} train a query rewriter based on the rewards relying on the ground-truth ranking list for information retrieval.
\citet{buck2018ask} use RL for single-turn question rewriting by maximizing the answers' quality which requires ground-truth. Similar to our F1 reward, \citet{wu2021conqrr} design rewards from ground-truth answers to train a conversational query rewriter. Instead, we propose alternative rewards indicating the confidence of answers from a model itself which do not require human annotations.

\section{Problem Definition}\label{sec:problem}
In CQA, each conversation contains a sequence of (question, answer) pairs $D=\{q_1,a_1,...,q_n,a_n\}$, where $a_i$ is the answer for question $q_i$. A conversational question $q_i$ can be ambiguous and its interpretation depends on the conversational context $c_i = \{q_1,a_1,...,q_{i-1},a_{i-1}\}$. 
The goal of CQR for QA is to learn a model $\mathcal{R}_\theta$, parameterized by $\theta$, that can translate $q_i$ associated with $c_i$ into $q_i^\prime$, so that the semantic meaning of $q_i^\prime$ is equivalent to  $q_i$.
\begin{equation}
    q_i^\prime = \mathcal{R}_\theta(q_i,c_i) \label{eq:rewrite}
\end{equation}

A pretrained single-turn QA model is expected to answer $q_i^\prime$ better than $q_i$. Note that the QA model can be trained from a single-turn dataset different from $D$ and is fixed when training the rewriter.
The motivation is to explore whether the already deployed single-turn QA models can be exploited to train a rewriter and reused without further training by accepting the rewritten questions. 

\section{Approach}\label{sec:method}

\subsection{Model Overview}\label{sec:overview}

\begin{figure}[t!]
\includegraphics[width=0.48\textwidth,clip,trim=5 10 0 5]{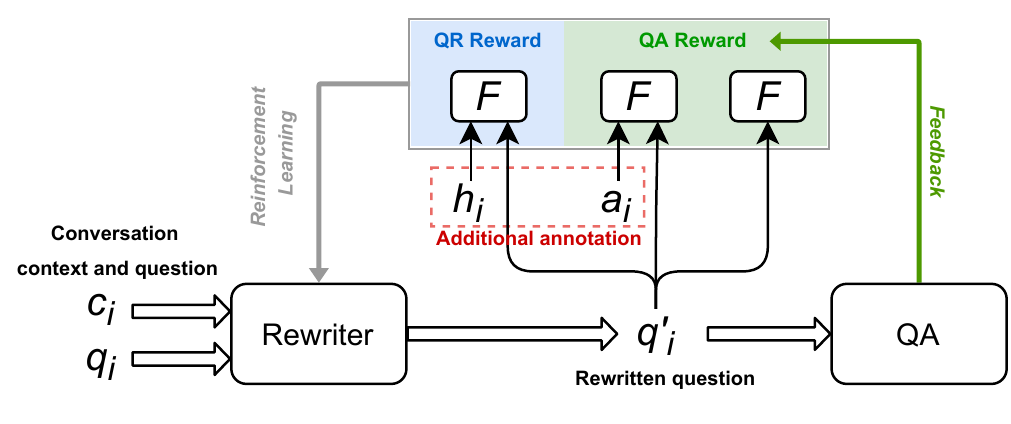}
\caption[]{Overview of our CQR approach. $h_i$ is human rewriting of $q_i$ and $a_i$ is the ground-truth answer of $q_i$. }
\label{fig:framework}
\end{figure}

We show our CQR approach with a modularized design in Figure \ref{fig:framework}. 
There are two major components: a CQR model $\mathcal{R}_\theta$ as introduced in Section \ref{sec:problem} and a reward function $\mathcal{F}$ that evaluates rewrite $q_i^{\prime}$ generated by $\mathcal{R}_\theta$ by producing a reward score. Then the CQR training can be formulated as a reinforcement training problem, where the objective is to maximize an expected reward or equivalently minimize the following loss function:
\begin{equation}\label{eq:exp}
    \mathcal{L}_{rl}(\theta) = - \mathbb{E}_{q_i^\prime \sim \mathcal{R}_\theta(q_i,c_i),\ q_i \sim \mathcal{T}}(\mathcal{F}(q_i^\prime))\ ,
\end{equation}
where $q_i$ comes from data distribution $\mathcal{T}$. 
During training, we push $\mathcal{R}_\theta$ to generate $q_i^\prime$ that achieves a higher reward by minimizing Equation~\ref{eq:exp}. Hereinafter, we omit $\theta$ from $\mathcal{R}_\theta$ for simplicity.

We define two types of rewards: QR rewards evaluate how similar a question rewrite is to the ground truth one produced by human annotators; QA rewards evaluate how well a QA model can answer a question rewrite. 
We summarize the characteristics of different rewards in Table~\ref{tab:rewards}. By maximizing one of the QR or QA rewards, we can explicitly optimize the model to achieve the QR or QA target.
Next, we describe the two types of rewards.

\begin{table}[th]
\centering
\resizebox{.99\columnwidth}{!}{%
\begin{tabular}{@{}ccccc@{}}
\toprule
\textbf{Reward} & \textbf{ROUGE} & \textbf{F1} & \textbf{Confidence} & \textbf{BM25} \\ \midrule
\textbf{Reward Type} & QR & QA & QA & QA \\
\textbf{CQA Type} & - & Extractive & Extractive & Retrieval \\
\textbf{Need Annotated Rewrites} & Y & N & N & N \\
\textbf{Need Annotated Answers} & N & Y & N & N \\ \bottomrule
\end{tabular}}
\caption{Characteristics of different rewards.}\label{tab:rewards}
\end{table}

\subsection{QR Rewards}
\label{sec:reward}
The rationale of maximizing QR rewards is similar to the aims of prior work: a good question rewrite should be similar to a human rewrite. We use the ROUGE-L score~\citep{lin-2004-rouge} between the question rewrite $q_i^\prime$ and the ground-truth $h_i$ as the QR reward:

\begin{equation}\label{eq:reward_rouge}
    \mathcal{F}(q_i^\prime,h_i) = ROUGE_L(q_i^\prime,h_i)
\end{equation}
This reward has been widely used by RL methods for language generation tasks. Note that Eq.~\ref{eq:reward_rouge} does not depend on the QA model and prior work can be considered as maximizing QR rewards. 

\subsection{QA Rewards}

We define QA rewards that reflect how well the question rewrites can help a QA model obtain better answers. Since QA rewards are task/model-dependent, we introduce QA rewards for the following two sub-types.

\subsubsection{Extractive CQA}

\textbf{Extractive CQA} is a machine reading comprehension (MRC) task and an extractive QA model $\mathcal{M}$  extracts the most likely span answer given a question $q$ and an evidence document $p$:
\begin{equation}
a_s =  \arg\max_{a_s} P_\mathcal{M}(a_s|q,p)    
\end{equation}

We assume that $\mathcal{M}$ is trained on regular single-turn QA data, and expects the input question $q$ to be self-contained. Therefore, CQA questions should be rewritten by $\mathcal{R}$ before being sent to $\mathcal{M}$. Next, we introduce supervised and unsupervised QA rewards.

\noindent\textbf{Supervised QA rewards.}
A straightforward way to measure the quality of a question rewrite $q_i^\prime$ in terms of QA is to calculate the similarity between the predicted answer by $\mathcal{M}$ with $q_i^\prime$ as input and the ground-truth answer $a_i$. We denote $a_s^\prime$ as the extracted answer span by $\mathcal{M}$ using the rewritten question $q_i^\prime$ as input. We measure the overlap between $a_s^\prime$ and $a_i$ by F1 score:

\begin{equation}\label{eq:f1}
    \mathcal{F}(q_i',a_i) = F1(\arg\max_{a_s^\prime} P_\mathcal{M}(a_s^\prime|q_i^\prime,p), a_i )
\end{equation}

Intuitively, the rewrite $q_i^\prime$ is better if $a_s^\prime$ is closer to the ground-truth answer. Compared with Equation~\ref{eq:reward_rouge}, Equation~\ref{eq:f1} depends on the ground-truth answers instead of human rewrites.

\noindent\textbf{Unsupervised QA rewards.}
For a predicted span $a_s^\prime$, $\mathcal{M}$ assigns a probability $r_c = P_\mathcal{M}(a_s^\prime|q_i^\prime,p)$ that reflects the model's confidence about the answer. We assume that a higher confidence score of an answer indicates that the QA model has a better question understanding.
Therefore, we directly use the probability of the most likely answer as the confidence reward for a question rewrite:
\begin{equation}\label{eq:conf}
    \mathcal{F}(q_i') = \max P_\mathcal{M}(a_s^\prime|q_i^\prime,p)    
\end{equation}

F1 rewards can be considered as judgment scores on predicted answers by humans since the ground-truth answers are used, while confidence rewards represent the model's self-judgments.

\subsubsection{Retrieval CQA} 
We also evaluate our method's generalization on a different \textbf{retrieval CQA} task, where the goal is to return a list of documents in descending order of relevance scores produced by a retrieval CQA model:

\begin{equation}
rel = \mathcal{M}(q,p)    
\end{equation}
where $p$ is a document. A retrieval CQA model usually consists of two stages. In the first stage, a lightweight ranking algorithm such as BM25 is used to retrieve top-k candidate documents. In the second stage, a more complex model such as BERT \citep{devlin-etal-2019-bert} is used to rerank candidate documents. 
Here, we use the BM25 score between a question and a document, which is a type of QA reward that does not use annotated answers:

\begin{equation}\label{eq:bm25}
    \mathcal{F}(q_i') = BM25(q_i^\prime,p)    
\end{equation}
We expect the rewrite $q_i'$ can retrieve documents with higher BM25 scores in the first stage than $q_i$ so that the performance in the re-ranking stage can also be improved. 

\subsection{Training}\label{sec:train}

There are two steps in our training framework. The {first step}, the pre-training step, which has the same supervised target as prior work. The objective is to minimize the cross-entropy loss between the model's prediction $q^{\prime}$ and human ground-truth rewrites $h$:
\begin{equation}\label{eq:sup_loss}
    \mathcal{L}_{sup}=-y_h\log y_{q'} \ ,
\end{equation}
where $y_h$ is the one-hot vector of $h$ and $y_{q'}$ is the distribution over tokens in $q^{\prime}$ predicted by the model. 
Supervised pre-training ensures the model has the basic ability to rewrite the original question given the conversational context. 

The second step continues training $\mathcal{R}$ with RL to maximize different rewards. In this work, we use Self-Critical Sequence Training (SCST) \citep{rennie2017self}. 
Given a question $q$, we generate two question rewrites $q^s$ and $q^\prime$. $q^s$ is generated by sampling the word distribution from $\mathcal{R}$ at each step, and $q^\prime$ is generated by $\mathcal{R}$ using greedy decoding. 
Then we minimize the following loss function:
\begin{equation}\label{eq:sc}
    \mathcal{L}_{rl} = (r^\prime-r^s) \sum_{t=1}^{N} \log P_{\mathcal{R}}(w_{t}^s|w_{1:t-1}^s,q,c)
\end{equation}
Here, $P_{\mathcal{R}}(\cdot)$, which is defined by $\mathcal{R}$, is the probability of generating the t-th word conditioning on previously generated tokens of $q_s$, the original question $q$ and conversational history $c$. \todo{add SC intuition} Intuitively, minimizing $\mathcal{L}_{rl}$ increases the likelihood of $q^s$ if it obtains a higher reward than $q^\prime$ (i.e. $r^s > r^\prime$), and thus maximizes the expected total reward. 
Given a reward function, we can obtain $r^\prime=\mathcal{F}(q^\prime)$ ($\mathcal{F}$ can be one of Equation~\ref{eq:reward_rouge},\ref{eq:f1},\ref{eq:conf},\ref{eq:bm25}) and $r^s=\mathcal{F}(q^s)$.

We only choose one of the reward functions to obtain the reward for a question.
We leave the combination of different rewards as future work.
Additional training procedure details are described in Appendix~\ref{algo}.

\begin{table*}[ht]
\centering
\resizebox{.65\linewidth}{!}{%
\begin{tabular}{@{}llccccc@{}}
\toprule
\multicolumn{1}{c}{\multirow{2}{*}{\textbf{QR Method}}} & \multicolumn{2}{c}{\textbf{QA Metrics}} & \multicolumn{4}{c}{\textbf{QR Metrics}} \\ \cmidrule(l){2-7} 
\multicolumn{1}{c}{} & \textbf{EM} & \textbf{F1} & \textbf{B-1} & \textbf{B-4} & \textbf{R-1} & \textbf{R-L} \\ \midrule
Human & 42.41 & 54.53 & - & - & - & - \\
Original & 38.41 & 48.95 & 61.06 & 30.98 & 69.91 & 69.71 \\
Co-reference & 38.17 & 48.99 & 54.95 & 30.84 & 74.11 & 73.40 \\
BART$_{CQR}$ & 41.26 & 53.60 & 64.20 & 39.33 & \textbf{76.70} & 74.00 \\ \midrule
RL-QR & 41.33 & 53.74 & \textbf{64.25} & \textbf{39.52} & \textbf{76.70} & \textbf{74.01} \\
RL-F1 & \textbf{41.91} & 54.27\textsuperscript{\textdagger} & 62.32\textsuperscript{\textdagger} & 37.79\textsuperscript{\textdagger} & 74.93\textsuperscript{\textdagger} & 72.09\textsuperscript{\textdagger} \\
RL-C & \textbf{41.91} & \textbf{54.61}\textsuperscript{\textdagger} & 57.47\textsuperscript{\textdagger} & 34.18\textsuperscript{\textdagger} & 71.12\textsuperscript{\textdagger} & 68.21\textsuperscript{\textdagger} \\ \bottomrule
\end{tabular}}
\caption{Overall QR and QA performance (\%) on CANARD. \textbf{Bold} indicates the best results except ``Human''. We denote BLEU-n as B-n and ROUGE-n as R-n. \textdagger\ denotes statistically significant difference from BART$_{CQR}$ ($p < 0.05$ with t-test). }\label{tab:extractive_QA}
\end{table*}

\begin{table*}[ht]
\centering
\resizebox{.75\linewidth}{!}{%
\begin{tabular}{llccccc}
\hline
\multicolumn{1}{c}{\multirow{2}{*}{\textbf{QR Method}}} & \multicolumn{2}{c|}{\textbf{QA Metrics}} & \multicolumn{4}{c}{\textbf{QR Metrics}} \\ \cline{2-7} 
\multicolumn{1}{c}{} & \multicolumn{1}{c}{\textbf{EM}} & \multicolumn{1}{c|}{\textbf{F1}} & \textbf{B-1} & \textbf{B-4} & \textbf{R-1} & \textbf{R-L} \\ \hline
BART$_{CQR}$ (50\%) & 41.37 & 53.52 & \textbf{63.83}& \textbf{38.88} & \textbf{76.57} & \textbf{73.79} \\
RL-C (50\%) & \textbf{42.09} & 54.76\textsuperscript{\textdagger} & 62.13 & 37.52 & 75.03 & 72.10 \\
RL-C (50\%+100\%) & 42.05 & \textbf{54.84}\textsuperscript{\textdagger} & 57.86\textsuperscript{\textdagger} & 34.44\textsuperscript{\textdagger} & 71.67\textsuperscript{\textdagger} & 68.54\textsuperscript{\textdagger} \\ \hline
\end{tabular}}
\caption{QR and QA performance (\%) of BART$_{CQR}$ and RL-C when using 50\% of ground-truth rewriting. \textdagger\ denotes statistically significant difference from BART$_{CQR}$ (50\%) ($p < 0.05$ with t-test).}
\label{tab:data_ratio}
\end{table*}

\section{Data and Experimental Setup}

\subsection{Datasets}

Similar to \citet{vakulenko2021question}, we experiment with CANARD \citep{elgohary-etal-2019-unpack} for extractive CQA and CAsT-19 \citep{Dalton2020TRECC2} for retrieval CQA. 
As CAsT-19 is small compared to CANARD, prior work \citep{vakulenko2021question} uses the same model trained on CANARD to evaluate the rewriting performance on the test set of CAsT-19. Similarly, we start with the modelnon CANARD, and continue RL training with the BM25 reward on the training set without using any human annotations provided by CAsT-19.

\subsection{Evaluation Metrics}

We use BLEU-1, BLEU-4, ROUGE-1 and ROUGE-L for automatic evaluation. We also evaluate the performance of rewrites on downstream QA tasks. For CANARD, we use F1 and Exact Match (EM).  For CAsT-19, we report MAP, MRR and NDCG@3 as in \citet{vakulenko2021question}.

\subsection{Baselines}

We consider the following baselines:

\noindent\textbf{Origin} uses the original conversational question as input of QA.

\noindent\textbf{BART$_{CQR}$} We fine-tune BART~\citep{lewis-etal-2020-bart} as a supervised baseline which has the same training procedure as the pre-training step of our method.

\noindent\textbf{Co-reference}~\citep{vakulenko2021question} is a rule-based method. We replace anaphoric expressions in original questions with their antecedents from the previous conversation turns. A public neural co-reference model~\citep{lee-etal-2018-higher} is used. 

\noindent\textbf{Human} uses the human rewrites and can be considered as an upper bound. However, we later show that the human baseline is the upper bound for QR target but not QA target.

\subsection{Implementation Details}

For all the QA models, we simulate the scenario where they are trained on single-turn QA data and cannot be updated when interacting with the rewriting component. The goal is to improve single-turn QA models for CQA, which means the input for QA models does not include any previous context.

\noindent{\textbf{Single-turn Extractive QA Model.}} 
To simulate a single-turn extractive QA model, we fine-tune ALBERT-XXLarge-v2 \citep{Lan2020ALBERT:} on the CANARD training set. 

\noindent{\textbf{Single-turn Retrieval QA Model.}} Same as in \citet{vakulenko2021question}, we use Anserini's implementation of BM25~\cite{robertson2009probabilistic} for the first-stage retrieval to obtain the top 1000 passages. In the second stage, we use BERT-large for passage re-ranking. Both components are fine-tuned on the MS MARCO dataset so that the two-stage pipeline resembles a single-turn retrieval QA model.  

\noindent{\textbf{Rewriting Models.}} Our RL-based methods and the supervised BART baseline (BART$_{CQR}$) use BART-base model~\cite{lewis-etal-2020-bart}.\footnote{The max sequence length is set to 284, with batch size 24. An Adam weight decay optimizer with an initial learning rate of 1e-5 is used to train those models for 10 epochs.} We use the official CANARD validation set for early stopping. \textbf{RL-QR} denotes the model when QR rewards are used. \textbf{RL-F1}, \textbf{RL-C} and \textbf{RL-BM25} denote models where the F1, confidence and BM25 rewards are used, respectively.

\begin{table*}[ht]
\centering
\resizebox{.70\linewidth}{!}{%
\begin{tabular}{lccccccc}
\toprule
\multirow{2}{*}{\textbf{QR Method}} & \multicolumn{3}{c|}{\textbf{QA Metrics}} & \multicolumn{4}{c}{\textbf{QR Metrics}} \\ \cmidrule(l){2-8} 
 & MAP & MRR & \multicolumn{1}{c|}{NDCG@3} & B-1 & B-4 & R-1 & R-L \\ \midrule
Origin & 17.85 & 46.44 & 27.86 & 71.63 & 51.54 & 82.65 & 81.24 \\
Human & 39.23 & 87.06 & 58.19 & - & - & - & - \\ \midrule
BART$_{CQR}$ & 28.02 & 61.49 & 44.04 & \textbf{75.12} & \textbf{55.54} & 84.82 & 83.84 \\
Co-reference & 26.82 & 59.74 & 43.05 & 71.19 & 51.79 & \textbf{88.06} & \textbf{87.69} \\
RL-BM25 & \textbf{28.41} & \textbf{63.20} & \textbf{45.54} & 71.92 & 52.01 & 82.92 & 81.59 \\ \bottomrule
\end{tabular}}
\caption{QR and retrieval performance (\%) on CAsT-19.}\label{tab:cast19}
\end{table*}

\section{Results}\label{sec:main_result}

Here, we study the following research questions:

    \noindent \textbf{RQ1}: Can our proposed QR and QA rewards improve the overall CQA performance? In particular, how effective are unsupervised rewards?
    
    \noindent \textbf{RQ2}: Does achieving the best QR target mean achieving the best QA target? %
    
    \noindent \textbf{RQ3}: What is the quality, as judged by humans, of the reward-guided question rewrites? %

\subsection{Evaluation on Extractive CQA}\label{rs:extractive}

We list the results on CANARD in Table~\ref{tab:extractive_QA}. EM and F1 are QA metrics while others are QR metrics. We observe several trends.

First, RL-based methods achieve the best results on both QA or QR metrics over other non-human baselines. Compared with BART$_{CQR}$, our proposed RL methods can further improve the performance on QA target and QR target. Specifically, RL-C outperforms BART$_{CQR}$ by 1.88\% and 1.58\% in terms of F1 and EM, respectively. RL-QR achieves marginally better scores on BLEU-1, BLEU-4 and ROUGE-L than BART$_{CQR}$.
RL-F1 achieves better F1 and EM scores than RL-QR and BART$_{CQR}$ but does not outperform RL-C. We notice that the F1 reward is less sensitive to question rewrites than the confidence reward. A small change in a question can lead to the same answer and F1 score. However, the confidence score can be different.
In this aspect, RL-C seems to differentiate the fluctuations on rewrites better than RL-F1. In answer to \textbf{RQ1}, the confidence reward is the most effective for CQA performance.  As an unsupervised reward which does not require either human rewrites or gold answers to a question, the confidence reward is even more effective than the F1 reward.  However, we do not claim or target state-of-the-art performance in our work. The goal is to verify whether our RL framework for CQR with different rewards can further improve the performance of a single-turn QA system for CQA.

Second, using QR rewards (RL-QR) leads to limited performance improvement under both QA and QR metrics compared with BART$_{CQR}$.  Maximizing the ROUGE rewards (Eq. \ref{eq:reward_rouge}) and minimizing the cross-entropy loss (Eq. \ref{eq:sup_loss}) share the similar intuition that a good reformulation from the model should be similar to human reformulated questions. 
The two\ objectives are very close and therefore lead to similar results.
It is important to note that the best scores of QR metrics and QA metrics are not achieved by the same method. Moreover, using QA rewards even lead to a large decrease in QR metrics.
Therefore, in response to \textbf{RQ2}, achieving the best QR target does not mean achieving the best QA target, and vice versa.

Third, \textbf{RL-C achieves higher F1 scores than the human baseline}.
Previous work (e.g.~\citealp{vakulenko2021question}) treats human annotations as an upper bound. However, we argue that more human-like rewrites do not guarantee better QA performance. 
The results verify our hypothesis that QA target does not necessarily align with QR target.
In \S\ref{sec:human}, we qualitatively analyze if rewrites generated by RL-C are better than the ground-truth.

\subsection{Training with Fewer Samples}\label{sec:ratio}

For a real-world CQA system, we can obtain a large number of user questions with no corresponding ground-truth rewrites or answers. Since the confidence reward can be obtained easily from the downstream QA models without requiring human annotations, we can use RL-C to continue training the rewriting model. We first train a baseline using $50\%$ of training data from CANARD (denoted as BART$_{CQR}$ (50\%) ). Then we continue RL training with the confidence reward using either the same $50\%$ data used in pre-training (denoted as RL-C (50\%)) or all questions in CANARD training set (denoted as RL-C (50\%+100\%)). 
The results are summarized in Table~\ref{tab:data_ratio}. We can see that RL-C (50\%+100\%) benefits from the large amount of questions during RL training and achieves better F1 and EM scores than RL-C (50\%). 
Interestingly, RL-C (50\%+100\%) outperforms the human baseline in Table~\ref{tab:extractive_QA} by 0.31\% in terms of F1. We also experimented with other ratios of data for supervised pre-training and continually RL training. In the experiments, we had similar observations that continual RL training with confidence rewards can further improve the downstream CQA performance.

\begin{table*}[ht]
\centering
\resizebox{.60\linewidth }{!}{
\begin{tabular}{@{}l|l|l@{}}
\toprule
 &  RL-C vs. BART$_{CQR}$ (\%) & RL-C vs. Human (\%)\\ \midrule
(1) RL is better & 121 (60.5\%) & 105 (52.5\%) \\
(2) RL is worse & 39 (19.5\%) & 58 (29.0\%) \\
(3) Both are good & 28 (14.0\%) & 33 (18.5\%)\\
(4) Both are bad & 12 (6.0\%) & 4 (2.0\%) \\
\hline
Total & 200 & 200 \\ \bottomrule
\end{tabular}}
\caption{Results of user study comparing two groups of rewrites using four preference options. }
\label{tab:human}
\end{table*}

\begin{table*}[ht]
\centering
\resizebox{0.78\linewidth}{!}{
\begin{tabular}{ccp{1.18\columnwidth}}
\hline
\multirow{4}{*}{Example 1} & Original & What happened after \textbf{he} was fired? \\
 & Human & What happened after \textbf{Aynsley Dunbar} was fired? \\
 & BART$_{CQR}$ & What happened after \textbf{Aynsley Dunbar} was fired? \\
 & RL-C & What happened after \textbf{Aynsley Dunbar} was fired \textbf{by Herbie Herbert in late 1978?} \\ \hline
\multirow{4}{*}{Example 2} & Original & What position did \textbf{he} play? \\
 & Human & What position did \textbf{Red Schoendienst} play? \\
 & BART$_{CQR}$ & What position did \uwave{Ernie} \textbf{Schoendienst} play? \\
 & RL-C & What position did \uwave{Don} \textbf{Schoendienst} play \textbf{in the Majors}? \\ \hline
\end{tabular}
}
\caption{Qualitative comparison of question rewrites. More examples are shown in Appendix~\ref{appen:examples}.}
\label{tab:cases}
\end{table*}

\subsection{Evaluation on Retrieval CQA}\label{rs:ir}

For RL-BM25, we use RL-C trained on CANARD as the pretrained model, then train it to maximize the BM25 reward, which can be readily obtained from the retrieval model. Results on CAsT-19 are shown in Table~\ref{tab:cast19}. As with extractive CQA, RL-BM25 achieves lower scores on QR metrics than baselines. However, it improves the NDCG@3 of BART$_{CQR}$ by relatively 3.4\%, which shows our framework also generalizes to retrieval CQA. Note that we do not use any supervised signals in CAsT-19 training set for RL training.

\subsection{Human Evaluation}\label{sec:human}

In addition to CQA performance, generating user-friendly rewrites is also important for real-world applications, since the rewrites sometimes will be displayed to users. To answer \textbf{RQ3}, we perform a user study to evaluate the quality of model generated rewrites.
Specifically, two groups are compared: 
(1) The first group contains the rewrites generated by RL-C and human rewrites; (2) The second group contains rewrites from RL-C and BART$_{CQR}$, respectively. For each group, we randomly choose 200 questions from CANARD testing set. For each pair, we collect human's judgments on which rewrite contains more accurate context and details from conversation history.

\looseness -1 The results are shown in Table~\ref{tab:human}. 
The study suggests that RL-C significantly performs better than Human and BART$_{CQR}$ (p-value < $0.001$, see details in Appendix~\ref{app_test}).
Remarkably, annotators prefer the rewrites from RL-C than humans in more than 50\% cases. We show two examples in Table~\ref{tab:cases}. In the first example, both RL-C and BART$_{CQR}$ correctly replace the pronoun with the referred person name. However, the rewrite generated by RL-C includes more accurate details which appear in conversation history. In the second example, both RL-C (same as RL-F1 and RL-QR) and BART$_{CQR}$ fail to generate the correct person's name. This error might be due to the prior knowledge of BART. 
To answer \textbf{RQ3}, we find that our reward-guided model can generate rewrites preferred by humans. However, all rewriting models can suffer from the coreference resolution problem.

\section{Conclusion}
We proposed a conversational question rewriting (CQR) approach using reinforcement learning. Such rewriting approaches are an emerging solution in real-world settings where QA systems with many existing answering backends trained on standalone questions must be adapted to work in conversational settings.

After assessing various QA and QR rewards, we showed that optimizing QR rewards is limited in improving CQA performance. In contrast, QA rewards that do not require ground-truth annotations consistently achieve the best CQA performance over baselines. For extractive CQA, using confidence rewards improved F1 by 2\% over BART-based baseline on CANARD; and for retrieval CQA, using BM25 rewards improved the NDCG@3 of the baseline by 3.4\% on CAsT-19. A human evaluation also demonstrated that our approach can generate higher-quality rewrites with more accurate and detailed context information.

\bibliography{ref}
\bibliographystyle{acl_natbib}

\clearpage
\appendix
\section*{\centering Appendix}

\section{Algorithm}\label{algo}
There are two steps for training the rewriting model. 
\begin{enumerate}
    \item The 1st step pre-training (line 1-6) is to minimize the cross-entropy loss between the model's prediction $q^{\prime}$ and human ground-truth rewrite $h$. This objective is used in most of prior work (e.g., \citet{vakulenko2021question}).
    \item The 2nd step (line 8-16) continues training the model with a reinforcement learning method (Self-Critical Sequence Training). In line 10 and 11, we only chose one of the reward functions to obtain the reward for a question. We leave the combination of different rewards as future work.
\end{enumerate}
\begin{algorithm}
  \small
     \SetKwInOut{Input}{Input}
     \SetKwInOut{Output}{Output}
     \Input{Initialized rewriter $\mathcal{R}$, human question rewrites $H$,
     conversations $\mathcal{T}_{pre}$ for pre-training,
     conversations $\mathcal{T}_{rl}$ for RL training, selected reward function $\mathcal{F}$}
    \Output{Trained rewriter $\mathcal{R}$}
    
    \tcc{Step 1: pre-training $\mathcal{R}$}\
    \For{$D\in\mathcal{T}_{pre}$} 
    {
        \For{ question $q$, context $c\in D$ and $h\in H$}
            {   generate $q^\prime = \mathcal{R}(q,c)$ (greedy decoding)\;
                minimize loss in Equation~\ref{eq:sup_loss}\;
                }
    }\
    \tcc{Step 2: self-critical training}\
    \For{$D\in\mathcal{T}_{rl}$}
    {
        \For{question $q$, context $c\in D$ and $h\in H$}
            {generate $q^s$ from $\mathcal{R}(q,c)$ by sampling\;
            generate $q^\prime$ from $\mathcal{R}(q,c)$ by greedy decoding\;
            obtain $r^s=\mathcal{F}(q^s)$\;
            obtain $r^\prime=\mathcal{F}(q^\prime)$\;
            minimize loss in Equation~\ref{eq:sc}\;
            }
    }
    \caption{CQR training}
\end{algorithm}

\section{Human Study Design}
For each annotation, an annotator is presented with the evidence document, conversation history, the original question and two rewrites. The annotator is required to select one from four options as listed in Table~\ref{tab:human}. The source of rewrite is anonymized. For each pair of rewrite, we randomly assign them to two options so that the judgments are not biased by the position of choices. We collect two judgments per rewrite pair. If there is a tie, we collect additional judgments. The final judgments are based on majority vote. 

\subsection{Appen Interface}
\begin{figure*}[t]
\centering
\includegraphics[width=0.8\textwidth]{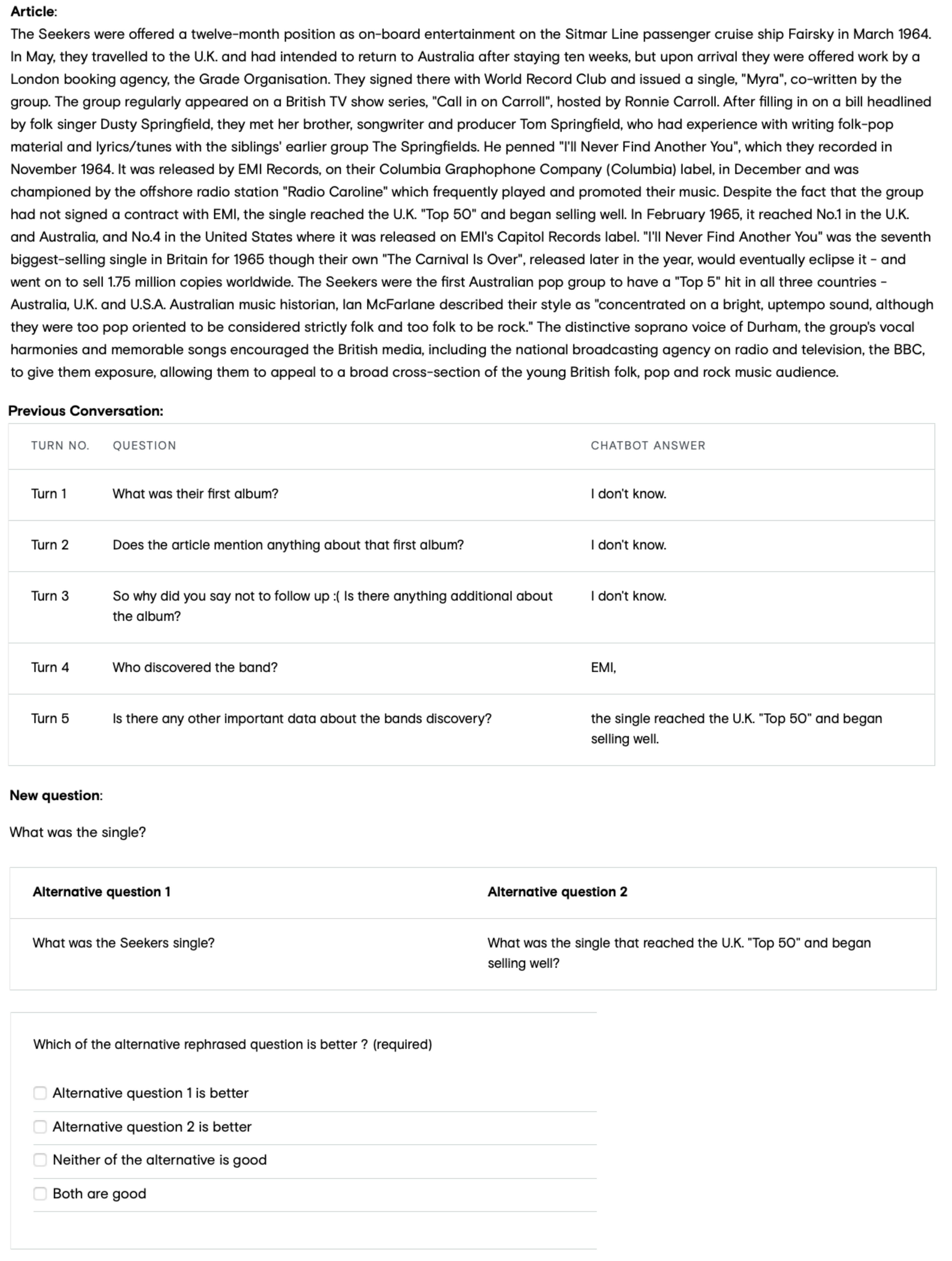}
\caption[]{Interface on Appen.}\label{fig:appen}
\end{figure*}

\begin{figure*}[t]
\centering
\includegraphics[width=0.8\textwidth]{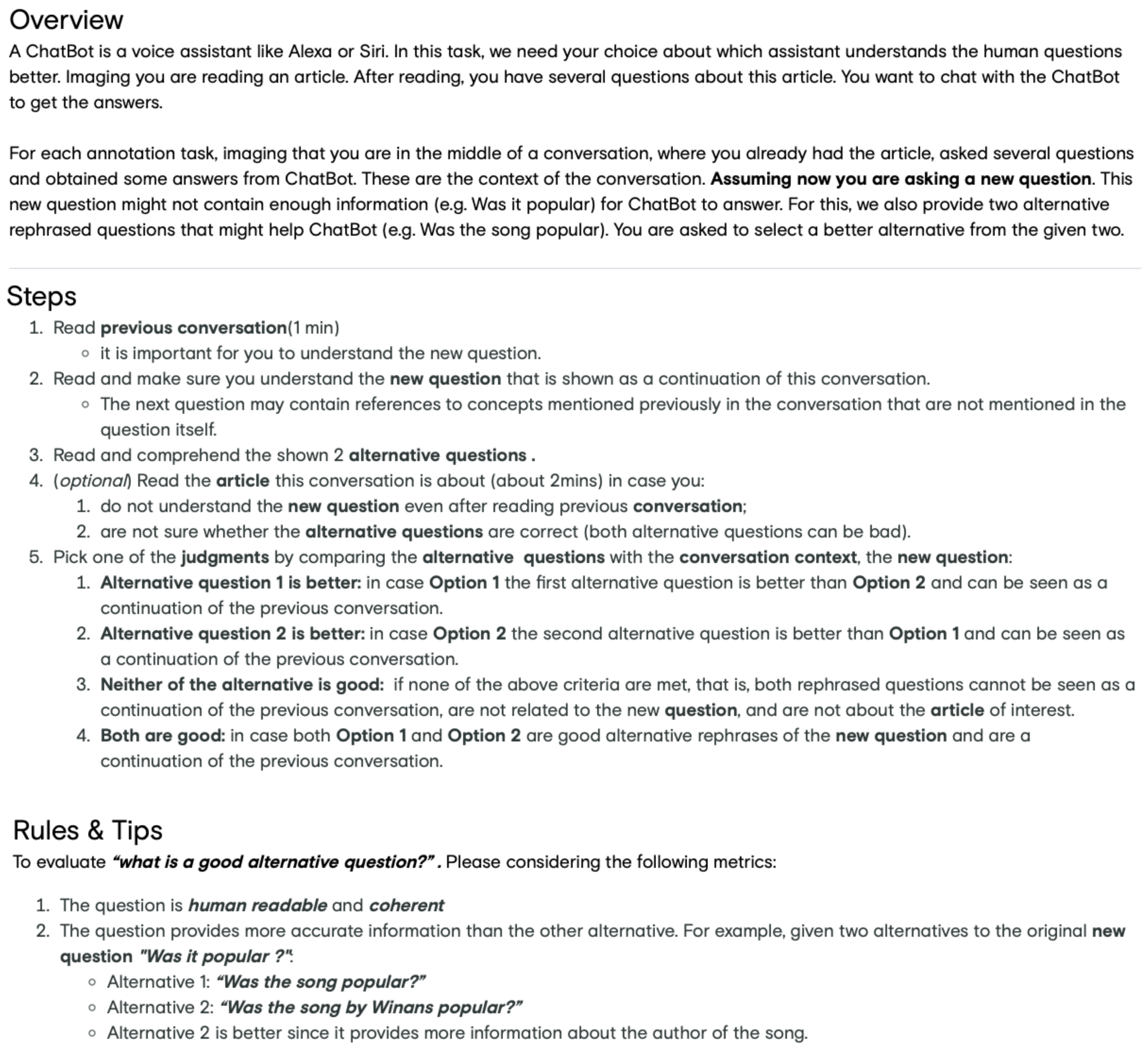}
\caption[]{Instruction for annotators. }\label{fig:instru}
\end{figure*}

\begin{figure*}[t]
\centering
\includegraphics[width=0.8\textwidth]{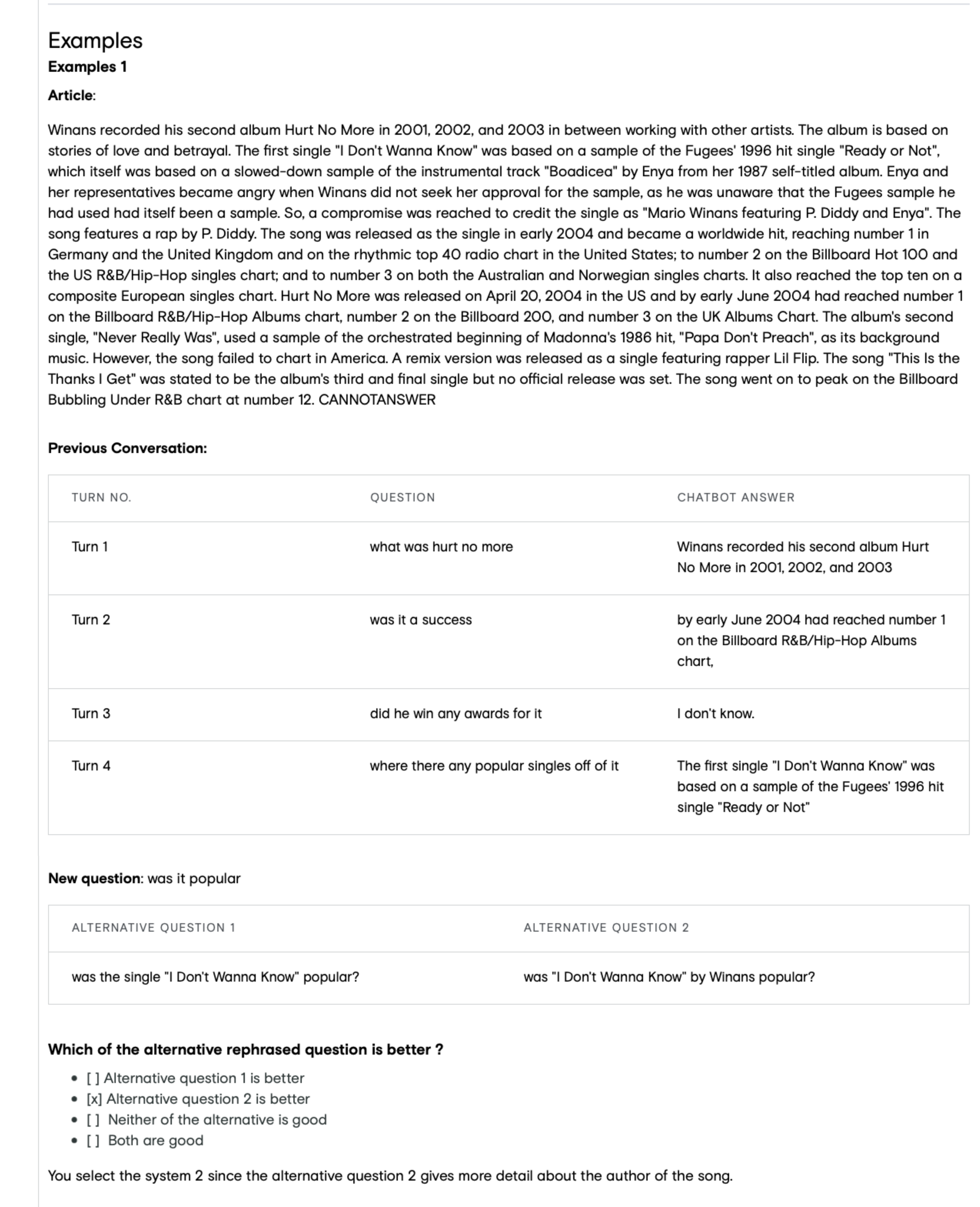}
\caption[]{An annotation example in the instruction. }\label{fig:appen_exp1}
\end{figure*}

Figure \ref{fig:appen} shows the interface for annotators. 
Figure \ref{fig:instru} contains the instruction which is visible for each annotator.
In the instruction, we show several annotation examples in Figure \ref{fig:appen_exp1}.

\subsection{Significance Tests} \label{app_test}
Here we describe how we conduct the Wilcoxon signed-rank test on the annotation results. When comparing RL-C with Human, for each sample, if annotators think RL-C is better, RL-C obtains score 1 and Human obtains score -1. Similarly, if annotators think Human is better, then Human obtains score 1 and RL-C obtains score -1. For other cases (i.e. both are good or both are bad), each of them obtains score 0. Then we use the method ``scipy.stats.wilcoxon'' in scipy library\footnote{https://docs.scipy.org} to do the test.
About the study annotator agreement rates, 48\% samples have 100\% agreement and the overall agreement rate is around 80\%.

\section{Rewriting Examples}\label{appen:examples}

In Table~\ref{appen:good:cases}, we show examples where the rewrites generated by RL-C are preferred by human annotators over the baseline method and ground truth. Compared with ground-truth rewrites, RL-C tends to generate rewrites with more factual details, which can help the user and also downstream QA systems to understand the question without conversation history. To some degree, it explains why the CQA performance is improved with RL-C, while the corresponding scores of QR metrics (i.e., BLEU-1, BLEU-4, ROUGE-1 and ROUGE-L) are very low. It also indicates that the human ground-truth in existing CQR datasets is not perfect and only evaluating CQR model with QR metrics can be biased.

The cases where both RL-C and the baseline generate incorrect rewrites are shown in Table~\ref{appen:bad:cases}. We can see that both methods make mistakes in coreference resolution. However, RL-C still has the tendency to include more conversational context in the rewrites.

\begin{table*}[t]
\centering
\resizebox{0.78\linewidth}{!}{
\begin{tabular}{ll}
\hline
Original & What kind of things   did \textbf{she} write about? \\
Human & What kind of things did \textbf{Le Guin} write   about? \\
BART$_{CQR}$ & What kind of things did \textbf{Le Guin} write   about? \\
RL-C & What kind of things did \textbf{Le Guin} write about \textbf{in Sociology}? \\ \hline
Original & What did \textbf{he} study at the university? \\
Human & What did \textbf{Chinua Achebe} study at the university? \\
BART$_{CQR}$ & What did \textbf{Chinua Achebe} study at the university? \\
RL-C & What did \textbf{Chinua Achebe} study at \textbf{the University of Ibadan}? \\ \hline
Original & What did \textbf{he} do after leaving Arrowverse? \\
Human & What did \textbf{John Barrowman} do after leaving Arrowverse? \\
BART$_{CQR}$ & What did \textbf{John Barrowman} do after leaving Arrowverse? \\
RL-C & What did \textbf{John Barrowman} do after leaving the Arrowverse \textbf{television franchise?} \\ \hline
Original & What kind of topics did \textbf{the show} cover? \\
Human & What kind of topics did \textbf{the Rush Limbaugh Show} cover? \\
BART$_{CQR}$ & What kind of topics did \textbf{Rush Limbaugh's show} cover? \\
RL-C & What kind of topics did \textbf{Rush Limbaugh's radio show} cover \textbf{in the 1970s}? \\ \hline
Original & What did he do after he landed? \\
Human & What did \textbf{Lindbergh} do after he landed? \\
BART$_{CQR}$ & What did \textbf{Charles Lindbergh} do after he landed \textbf{at Le Bourget Aerodrome?} \\
RL-C & \begin{tabular}[c]{@{}l@{}}What did \textbf{Charles Lindbergh} do after he landed \textbf{at Le Bourget Aerodrome} \\ \textbf{at 10:22 p.m. on Saturday, May 21, 1927}?\end{tabular} \\ \hline
\end{tabular}}
\caption{Examples of rewrites where the reformulated questions from RL-C are the best judged by human annotators.}
\label{appen:good:cases}
\end{table*}

\begin{table*}[t]
\centering
\resizebox{0.78\linewidth}{!}{
\begin{tabular}{ll}
\hline
Original & Did others agree with \textbf{him}? \\
Human & \begin{tabular}[c]{@{}l@{}}Did others agree with \textbf{Gottfried Wilhelm} on the idea that the truth of   religion and philosophy \\ cannot contradict with each other?\end{tabular} \\
BART$_{CQR}$ & Did others agree with \uwave{Leibniz's Theodicy?} \\
RL-C & \begin{tabular}[c]{@{}l@{}}Did others agree with \uwave{Leibniz}'s ideas that the truths of theology   (religion) and philosophy \\ cannot contradict each other,  since reason and faith are both "gifts of God" in the Theodicy?\end{tabular} \\ \hline
Original & What did \textbf{he} rejoin as? \\
Human & What did \textbf{Eddie Collins} rejoin as? \\
BART$_{CQR}$ & What did \uwave{Chris Hedges} rejoin as? \\
RL-C & What did \uwave{Chris Hedges} rejoin as in August? \\ \hline
Original & What year did \textbf{his} first film debut? \\
Human & What year did was \textbf{Paul Verhoeven}'s first film debut? \\
BART$_{CQR}$ & What year did \uwave{Steven Seagal}'s first film debut? \\
RL-C & What year did \uwave{James Cameron}'s first film debut? Flesh and Blood (1985)? \\ \hline
Original & Did \textbf{he} go into acting then? \\
Human & Did \textbf{Coogan} go into acting after college? \\
BART$_{CQR}$ & Did \uwave{Charlie Chaplin} go into acting after \uwave{A Day's Pleasure}? \\
RL-C & Did \uwave{Charlie Chaplin} go into acting after college? \\ \hline
Original & Did \textbf{they} do a second album? \\
Human & Did \textbf{Gerry Mulligan and Chet Baker's quartet} do a second album? \\
BART$_{CQR}$ & Did \uwave{Pacific Jazz} do a second album? \\
RL-C & Did \uwave{Pacific Jazz} do a second album \textbf{after PJ-8?} \\ \hline
\end{tabular}}
\caption{Examples question rewrites where both RL-C and BART$_{CQR}$ make mistakes.}
\label{appen:bad:cases}
\end{table*}

\end{document}